\documentclass[letterpaper, 10 pt, conference]{ieeeconf}  

\IEEEoverridecommandlockouts                              

\usepackage{graphicx}
\usepackage{amsmath}
\usepackage{amssymb}
\usepackage{xcolor}
\usepackage{booktabs}
\usepackage{subcaption}

\title{\bf CLARA: A Cognitively-Driven Clarification Framework for Grounding Ambiguous Requests}

\author{Dimitrios Bekris$^{1}$, Marios Glytsos$^{1}$, Panagiotis Filntisis$^{1}$, Niki Efthymiou$^{1}$, \\ George Retsinas$^{1}$ and Petros Maragos$^{2}$
\thanks{$^{1}$
D. Bekris, M. Glytsos, P.P. Filntisis, N. Efthymiou and G. Retsinas are with the Robotics Institute, Athena Research Center, 15125 Maroussi, Athens, Greece
{\tt\footnotesize d.bekris@athenarc.gr, m.glytsos@athenarc.gr, pfilntisis@athenarc.gr, nefthymiou@athenarc.gr, george.retsinas@athenarc.gr}
}
\thanks{$^{2}$
P. Maragos is with the School of ECE, NTUA, Athens, Greece, the Robotics Institute, Athena Research Center, Athens, Greece and the HERON - Hellenic Robotics Center of Excellence, Athens, Greece 
{\tt\footnotesize maragos@cs.ntua.gr}
}
\thanks{This work was carried out in early 2025; all comparisons reflect
the state of LVLMs and image-generation models available at that time.}
}

\begin{document}

\maketitle
\thispagestyle{empty}
\pagestyle{empty}

\begin{abstract}
Robotics and AI systems are increasingly deploying large vision-language models (LVLMs) to follow high-level instructions, from routine household chores to more complex organizational tasks. Yet, broad directives like “clean the room” or “organize the desk” frequently omit crucial details about user preferences and specific arrangement of objects that influence the desired outcome. Without any further guidance, such models tend to overestimate their grasp of ambiguous instructions and frequently assume they have sufficient understanding despite missing important details. In this paper, we propose CLARA (Conversational cLArification for Robotic Arrangement), a cognitively-driven clarification framework that proactively elicits key information from users, in a similar fashion as humans' clarification seeking behavior. Through targeted questioning, CLARA refines ambiguous requests into structured, detailed instructions describing precisely how objects should be organized or rearranged, according to the user's preferences. These clarifications then guide a diffusion-based image generation pipeline, producing realistic visual renderings of the final, desired scene. This final rendering also serves as a visual confirmation of the system's understanding from the user. Experimental results and user studies demonstrate that CLARA leads to more accurate and consistent outcomes compared to another State-of-the-Art LVLM system, highlighting the importance of interactive clarification for bridging the gap between vague directives and fully grounded task specifications.
\end{abstract}

\section{Introduction}


Robotics and intelligent systems/agents increasingly rely on large vision-language models (LVLMs) \cite{zhang2024vision,Qwen-VL, liu2023llava,liu2023improvedllava,lu2024deepseekvl} to interpret and execute human instructions, in an effort to simplify tasks ranging from household chores to complex workflows\cite{kannan2024smart,wang2024large, zeng2023large,gao2024physically}. Crucial to their effectiveness is accurately grounding ambiguous high-level commands, such as ``clean the room" into precise visual outcomes according to the desires of the instructor, since this can mean drastically different things based on an individual’s expectations, cultural norms, or personal preferences.  Unlike structured tasks in controlled environments, everyday scenarios often involve vague instructions and considerable variation in how individuals imagine the outcomes, making the task of LVLMs more difficult to reliably translate user intent from limited textual input alone.

From a cognitive standpoint, clarification-seeking behavior is recognized as a fundamental part of how humans manage ambiguous information. Conversation is viewed as a cooperative, joint activity in which speakers and listeners work together to ensure mutual understanding (i.e., establishing ``common ground"). Developmentally also, adults often scaffold children’s understanding by prompting them to explain or by clarifying instructions (since initially children tend to overestimate their understanding of an ambiguous message), which over time teaches children how to resolve ambiguities through conversation. Interestingly, LVLMs as well tend to in many cases overestimate their understanding of the user's query, and do not engage in dialogue to refine their understanding. In this paper, inspired by such clarification-seeking behavior, we aim to equip LVLM-powered robotic agents with a proactive questioning mechanism designed to clarify the user's intent.

Most existing grounding methods for vision-language models depend on single-turn, prompt-based interactions, assuming initial instructions fully represent user intentions \cite{brooks2022instructpix2pix, minidalle3}. While those methods are effective for simple direction, they frequently struggle with high level ambiguous requests. For instance, an instruction like "clean the room" can lead to diverse visual interpretations based on different individual expectations regarding organization or aesthetics. Although recent diffusion-based image generation models \cite{rombach2021highresolution, minidalle3} have significantly improved prompt to image translation, they rely on explicitly detailed prompts therefore limiting the applicability where user intentions remain underspecified. More adaptive approaches incorporate conversational grounding, allowing users iterative clarification through dialogue \cite{dialogpaint, Cui2023ChatEditTM, nguyen2024instruction}. Yet, current methods mostly allow small, step by step improvements without actively asking users for important clarifications, making them less useful in real situations.

To overcome these limitations, in this paper, we propose a framework that enables robotic agents to actively ask targeted questions, effectively grounding and clarifying abstract user goals like ``clean this room", converting them into specific, visually coherent outcomes. More specifically, we propose a chatbot-based framework designed to ground ambiguous user goals through targeted questioning and iterative interaction. Our approach initiates a guided dialogue, proactively asking clarifying questions in an effort to capture the user's precise visual intent. Leveraging these input clarifications, our system generates a visualization of the final scene, in order to verify with the user that it correctly understood their intentions.


In summary, our key contributions are:
\begin{itemize}
\item A conversational grounding framework inspired by human clarification-seeking behavior, that proactively engages users through targeted clarifying questions to accurately determine visual intent.
\item Integration of the interactive dialogue with image editing and diffusion-based image generation to better align the user's intentions(goals) with the system's ``understanding", and verify the system's clear understanding of the instruction.
\item A comparative user study with a a state-of-the-art LVLM-based system (Gemini 2.0), which shows that LVLMs augmented with such clarification seeking behavior are significantly preferred by the users in terms of their effectiveness for task completion.
\end{itemize}

\section{Related Work}

\textbf{Grounding in Vision-Language Models} Vision-language grounding methods typically focus on establishing accurate correspondences between textual instructions and visual representations. Early methods, such as Visual Dialog \cite{Das_2017_CVPR} and GuessWhat?! \cite{guesswhat_game}, established conversational tasks requiring agents to interpret user questions grounded explicitly in visual scenes. However, these primarily tested understanding rather than sicking clarification or iterative goal alignment.

Recent developments in large vision-language models (LVLMs) such as LLaVA \cite{liu2023llava, liu2023improvedllava} have enhanced open-ended dialogue about images, yet their interactions remain largely user-driven, without the system clarifying ambiguous commands. Models like ChatterBox \cite{tian2024chatterbox} have advanced multi-turn conversational grounding by refining spatial references through dialogue  but still emphasize reference resolution e.g. identification and location of objects through rather than task and goal oriented visual manipulation.

\textbf{Clarifying Ambiguous Instructions} Effective communication between human and machines often involves clarifying ambiguous instructions through dialogue, a challenge increasingly recognized in human-AI interaction research. Min et al. \cite{min2024situatedinstructionfollowing} highlight the need for robotic agents to handle underspecified instructions through implicit inference or explicit questioning, highlighting that actively clarifying user intentions through interaction is essential for accurately capturing and achieving their desired outcomes. Selective clarification approaches like CLAM \cite{kuhnCLAM2022} demonstrate decision making based on uncertainty estimates but primarily address linguistic ambiguity rather than visual task grounding.
Despite recognizing the value of clarification, current LVLMs rarely autonomously initiate clarification questions and tend to overestimate their information for completing tasks. Our work directly addresses this gap by enabling the chatbot to proactively query the user in order to clarify ambiguous instructions into precise visual outcomes.

Recent conversational editing approaches, including DialogPaint \cite{dialogpaint} and ChatEdit \cite{Cui2023ChatEditTM}, address iterative refinement through dialogue but mainly focus on incremental adjustments after initial generation, without proactive clarification. These methods iteratively apply image edits based on sequential user instructions, typically building upon previous outputs. Consequently, minor inaccuracies introduced in early stages may compound over successive interactions, gradually diverging from the user's original intent. 



\textbf{Multimodal Planning and Goal Decomposition} Recent multimodal planning frameworks, such as HuggingGPT \cite{shen2023hugginggpt} and GenArtist \cite{wang2024div}, demonstrate the effectiveness of breaking complex visual tasks into manageable sub-goals orchestrated by language models. HuggingGPT leverages LLMs for task decomposition and multi-step execution across diverse models, while GenArtist utilizes iterative verification to ensure accurate task execution.

Although these planning frameworks excel at structured decomposition, they typically rely on automated inference without interactive user clarification. Embodied agents like SayCan \cite{saycan2022arxiv} introduce physical grounding through feasibility checks but similarly lack explicit conversational clarifications from the user.



\section{Method}
\label{sec:method}

\begin{figure*}[t]
    \centering
    \includegraphics[width=1.0\textwidth]{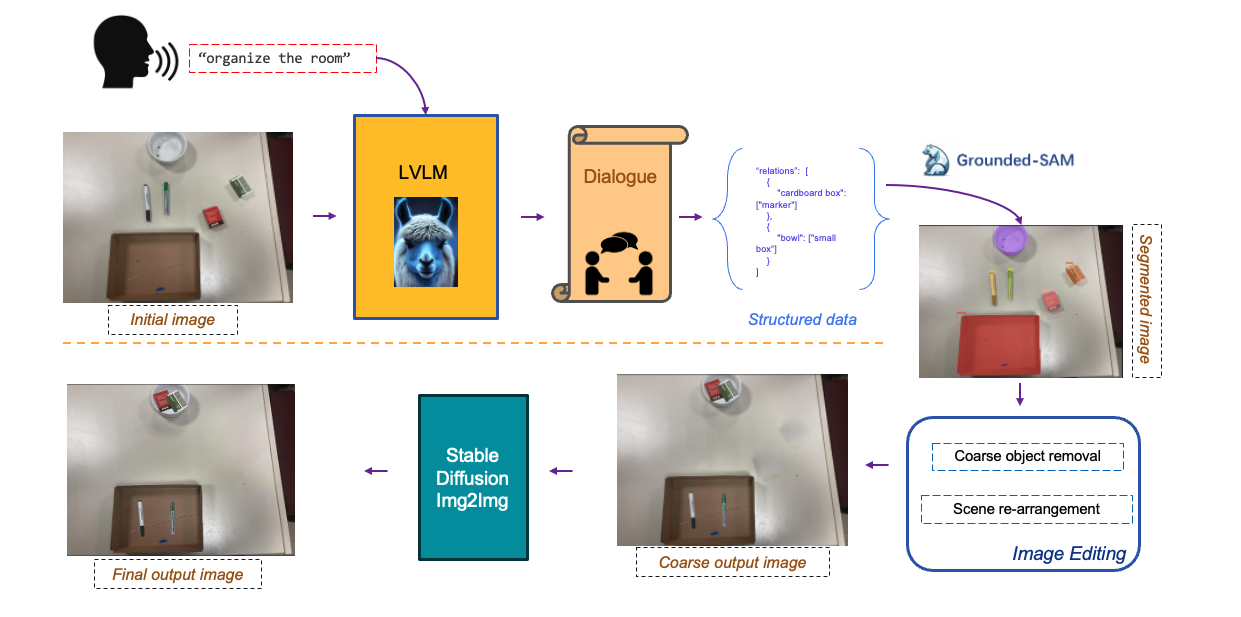}
    \caption{CLARA system pipeline. Given an input image and a high-level goal, an LVLM-based chatbot refines the user’s intentions via clarifying questions. The refined specification then guides scene segmentation, object rearrangement, inpainting, and final image generation.}
    \label{fig:pipeline}
\end{figure*}

The goal of CLARA is to align abstract user instructions, such as ``organise this room", with expected visual outcomes by clarifying the user’s intent. After clarifying the desired structure, the system proceeds to generate an image of the final scene, in order to verify that it understood the user's instructions. The pipeline of the system can be seen in Figure~\ref{fig:pipeline}. We first provide the high level overview of the system and subsequently describe each stage in detail.


\subsection{System Overview}

\textbf{Input and Goal Specification.} The user provides a single image depicting the initial scene (e.g., a room) together with a concise, abstract instruction (e.g., “clean the room”). Since such commands are inherently ambiguous, the system must uncover the user’s actual preferences for object placement, storage, and arrangement.

\textbf{Conversational Clarification.} Leveraging a large vision-language model (LVLM), our system engages with the user to clarify how they want to transform the scene (Sect.~\ref{subsec:chatbot}). For example, if the user simply says ``organize the desk", the system might ask: 'Should books be placed on the shelf, or remain stacked on the table?'

\textbf{Visual Generation and Refinement.} After confirming that it has sufficiently clarified the initial vague instruction, the system then translates these refined directives into a final image. This process involves segmenting and repositioning objects, discarding any specified items, and employing a generative model to achieve realistic completions. (Sec.~\ref{subsec:image_generation}).

\subsection{Conversational Clarification via Chatbot}
\label{subsec:chatbot}

To handle abstract or underspecified commands, we employ a custom LVLM-powered chatbot based on \textbf{LLAMA~3.2 (11B)}\cite{grattafiori2024llama}. Unlike typical single-turn prompt-based systems, our chatbot actively asks targeted clarification questions. It is initialized with guidelines instructing it to:

\begin{itemize}
    \item \textbf{Focus on spatial relations.} Elicit which objects the user wants moved, stored, or grouped together.
    \item \textbf{Confirm containers and storage details.} Pinpoint which items go into specific containers, plates, bins, etc.
    \item \textbf{Stop once sufficient clarity is achieved.} Upon judging that it has enough information, it provides a concise textual description of the final desired state, ensuring no important details are missing.
\end{itemize}

This way, the system can transform an initial vague request (e.g., “clean the room”) into a structured description specifying how each object should be handled. For example:
\begin{quote}
\emph{“Books: move to left shelf. Place all pens and notebooks in the top drawer. Clear trash from floor into bin.”}
\end{quote}
This clarifying dialogue significantly reduces ambiguity and sets the stage for visually accurate scene editing which happens in the next stage.

\subsection{Visual Generation and Refinement}
\label{subsec:image_generation}

After generating the textual structured description, the system generates an updated image illustrating the requested changes. This step helps verify that the system has successfully clarified the desires of the user.  

The visual generation and refinement stage follows the next steps:

\paragraph{Grounded Segmentation.}
To identify and segment the individual objects within the original scene based on their textual description, we adopt \textbf{Grounded-SAM}, an integration of Grounding DINO~\cite{liu2023grounding} and Segment Anything Model (SAM)~\cite{kirillov2023segany}. Grounding DINO provides open-set object detection (i.e., it can localize arbitrary objects given textual queries), while SAM refines those detections into precise segmentation masks. 

\paragraph{Scene Rearrangement.}
Now, using the structured directives produced by the Chatbot (e.g., “move the lamp to the corner table”), each identified object is (automatically) virtually repositioned or removed. In cases where the user wants an object hidden (e.g., “throw away all crumpled papers”), we remove the relevant masks from the original image.

\paragraph{Coarse Inpainting.}
Because the previous object removal or movement leaves visual gaps in the original image, we initially fill these voids with \textbf{LaMa inpainting}~\cite{suvorov2021resolution}, a specialized model for reconstructing missing regions. Although LaMa produces coherent completions, the results can appear slightly disjointed or lack the full realism often expected in complex scenes.

\paragraph{Stable Diffusion Refinement.}
To achieve a more photorealistic and harmonious final image, we pass the inpainted output through a \textbf{Stable Diffusion} model~\cite{rombach2021highresolution}, conditioning on the user’s clarified description. By conditioning on both textual guidance (e.g., “the desk is free of clutter, with the books neatly on the shelf”) and the partially inpainted image, Stable Diffusion refines object boundaries, ensures consistent lighting, and enhances overall realism. This conditioning step allows the model to preserve crucial details of the scene layout while seamlessly integrating any newly inpainted or rearranged content.

Finally, the system presents the rendered scene to the user for verification. If the user indicates that the arrangement does not fully capture their desires or requests further modifications, the chatbot re-engages to clarify any outstanding ambiguities. This iterative process can continue until the user confirms that the final image accurately reflects their intended outcome.

\paragraph{Verification.} Finally, the system presents the rendered scene to the user for evaluation. This ensures that the user can confirm whether the final image accurately reflects their intended outcome.


\section{Results and Evaluation}
\label{sec:results}

\subsection{Experimental Setup}

\paragraph{Dataset and Prompts}
We designed three distinct abstract scenarios involving common household settings, each associated with ambiguous, high-level instructions. The scenarios include:

\textbf{Scenario 1: Organize a Desk.} The model is provided with an image of an unorganized desk with various items and is asked to \textit{"tidy the desk up"}.

\textbf{Scenario 2: Serving food into platters.} The model is provided with an image of a dinner table with fruits and platters and is asked to \textit{"serve the food"}.

\textbf{Scenario 3: Clean a desk.} The model is provided with an image of a dirty desk with trash and other various items and is asked to \textit{"clean it."} 

\begin{figure}
    \centering
    \includegraphics[width=0.32\linewidth]{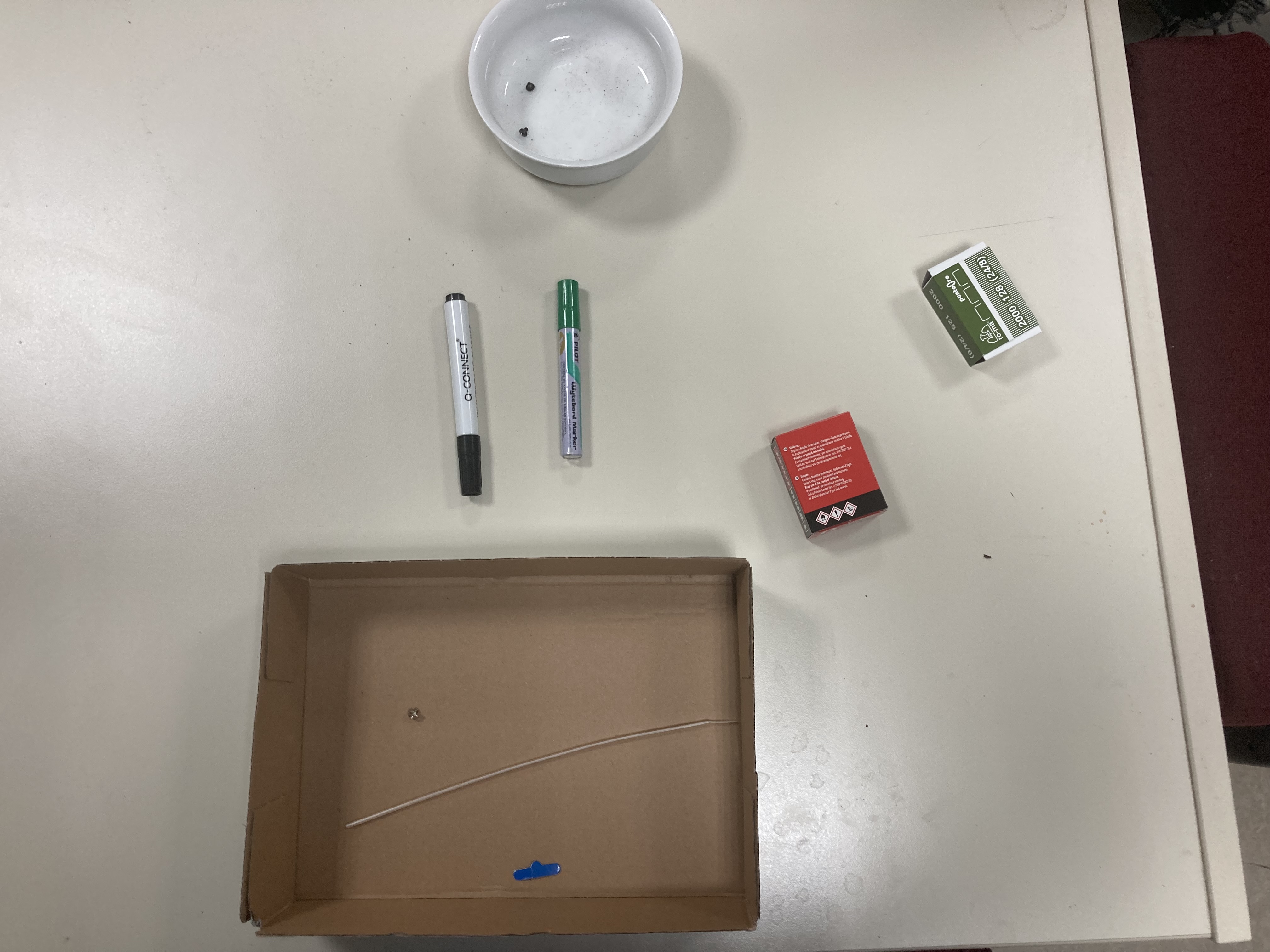}
    \includegraphics[width=0.32\linewidth]{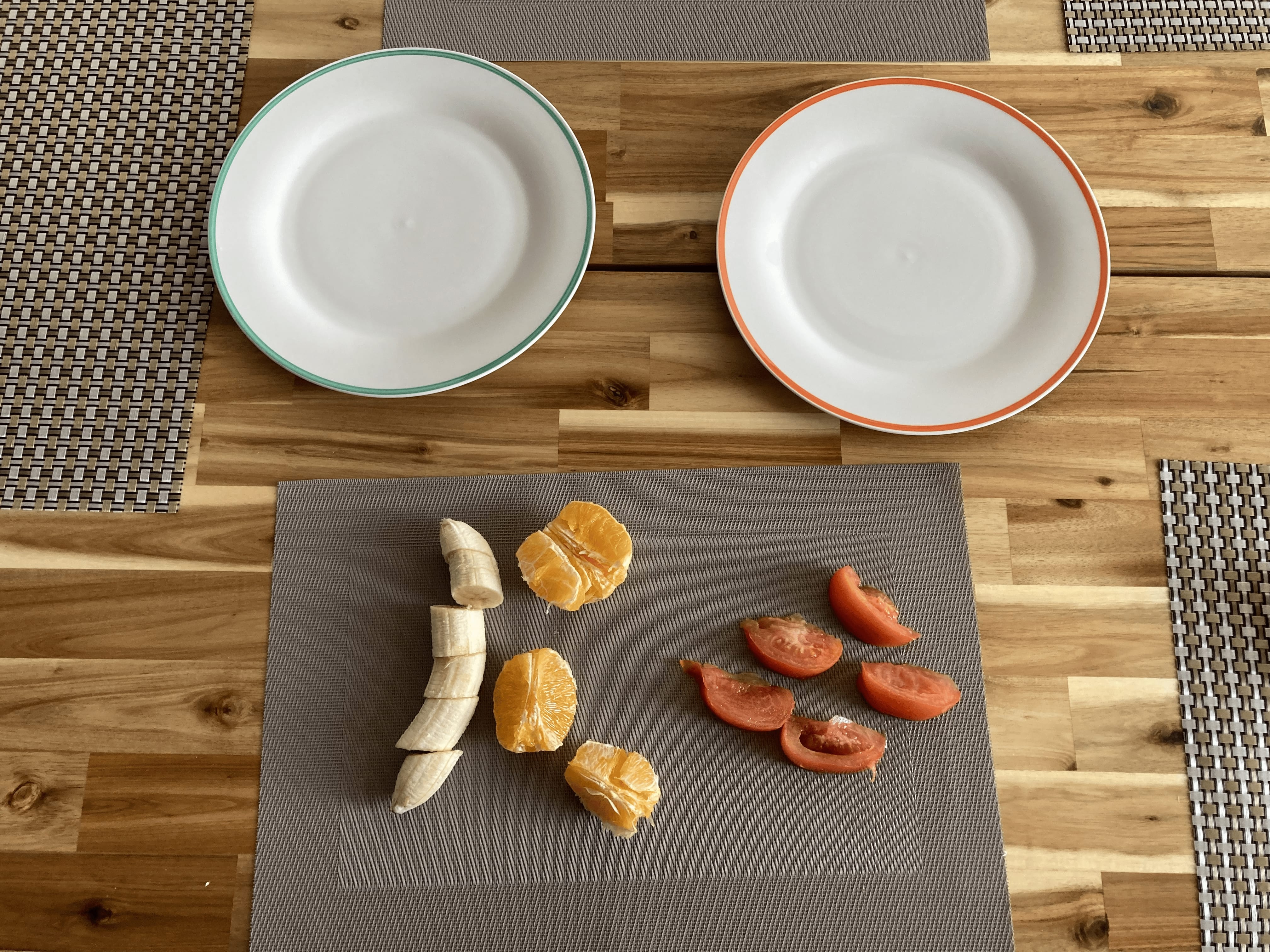}
    \includegraphics[width=0.32\linewidth]{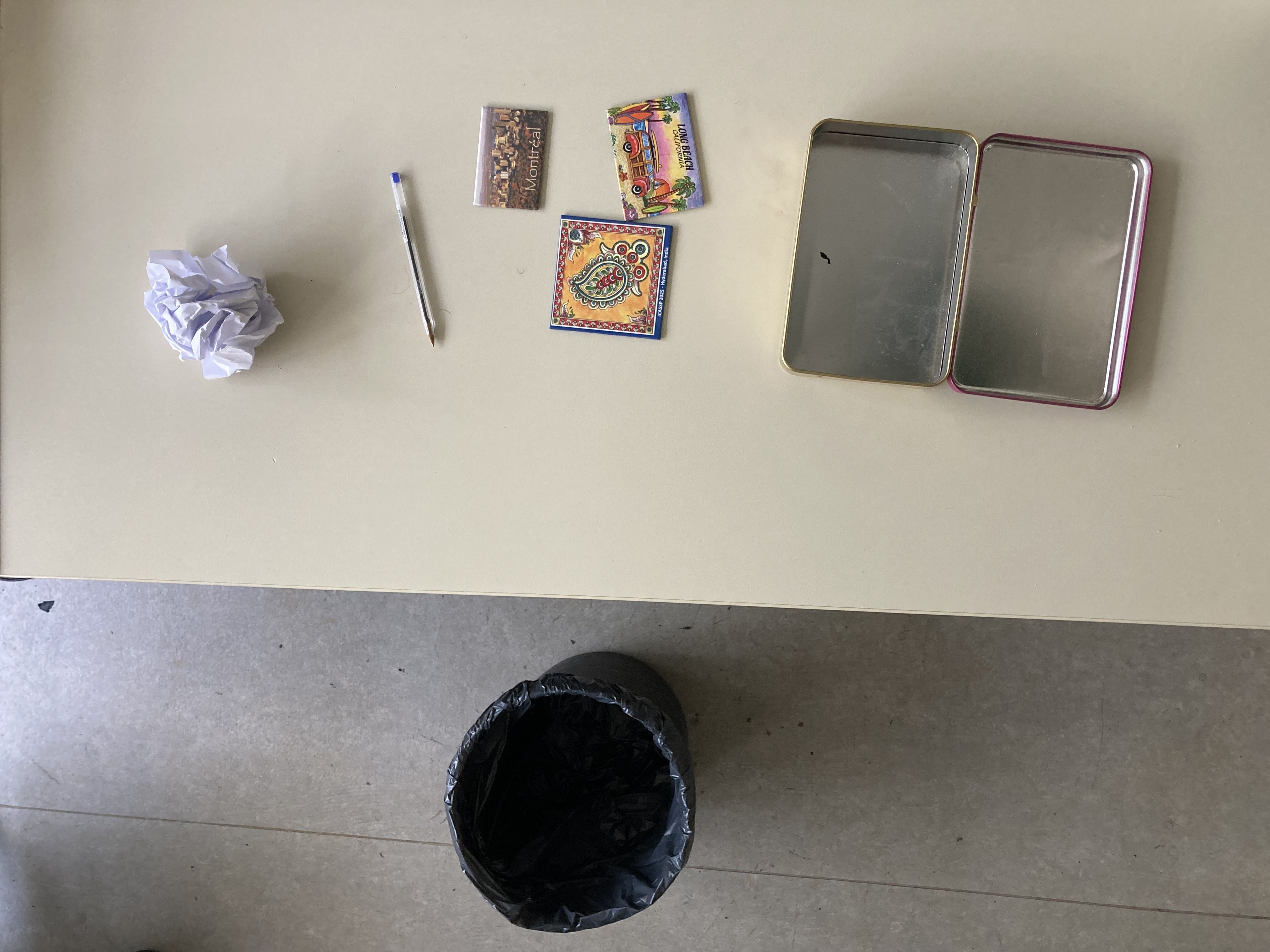}
    \caption{The abstract scenarios designed for the experimental setup. An unorganized desk (left), a dinner table with food to serve (middle), and a dirty desk (right).}
    \label{fig:scenarios}
\end{figure}

The three images of these scenarios can be seen in Figure~\ref{fig:scenarios}.

\paragraph{Methods}
We assess the effectiveness of CLARA by comparing it against \textbf{Gemini~2.0 Flash (Experimental)}~\cite{geminiteam2024geminifamilyhighlycapable}, a next-generation multimodal model from Google AI that is also capable of generating images. We also perform an ablation study where we combine the conversational clarification part of CLARA with the image generation capabilities of Gemini (i.e., we feed Gemini with the final clarified directives extracted by CLARA). We call this model CLARA+Gemini.

\subsection{Quantitative and Qualitative results}
In order to quantitively assess the performance of each system to successfully clarify the abstract instruction of the user we run for each system and each scenario 10 times \textit{using the same prompt}, summing up to a total of 30 comparisons per model. More specifically, we ran \emph{Gemini} in single-step mode: for each prompt (e.g., “Organize this desk.”), Gemini generated 10 images by varying the random seed. For CLARA, we conducted a single interactive dialogue per prompt, asking for the same object arrangement. Once finalized, we generated \emph{one} final image from our pipeline.

We report two main quantitative measures:

\paragraph{Standard Deviation in Embedding Space.}
We extracted an image embedding for each output using CLIP~\cite{clip} and measured the standard deviation across the 10 generations. Lower values indicate more consistent outputs. As shown in Table \ref{tab:std}, CLARA's outputs exhibit lower standard deviation. 

\paragraph{Cosine Similarity}
Similarly, we measured the mean value of pairwise cosine similarity among the embeddings. Here, higher values indicate more consistent outputs. As shown in Table \ref{tab:std}, CLARA's outputs exhibit higher cosine similarity.

\begin{figure}
    \centering
    \includegraphics[width=1.0\linewidth]{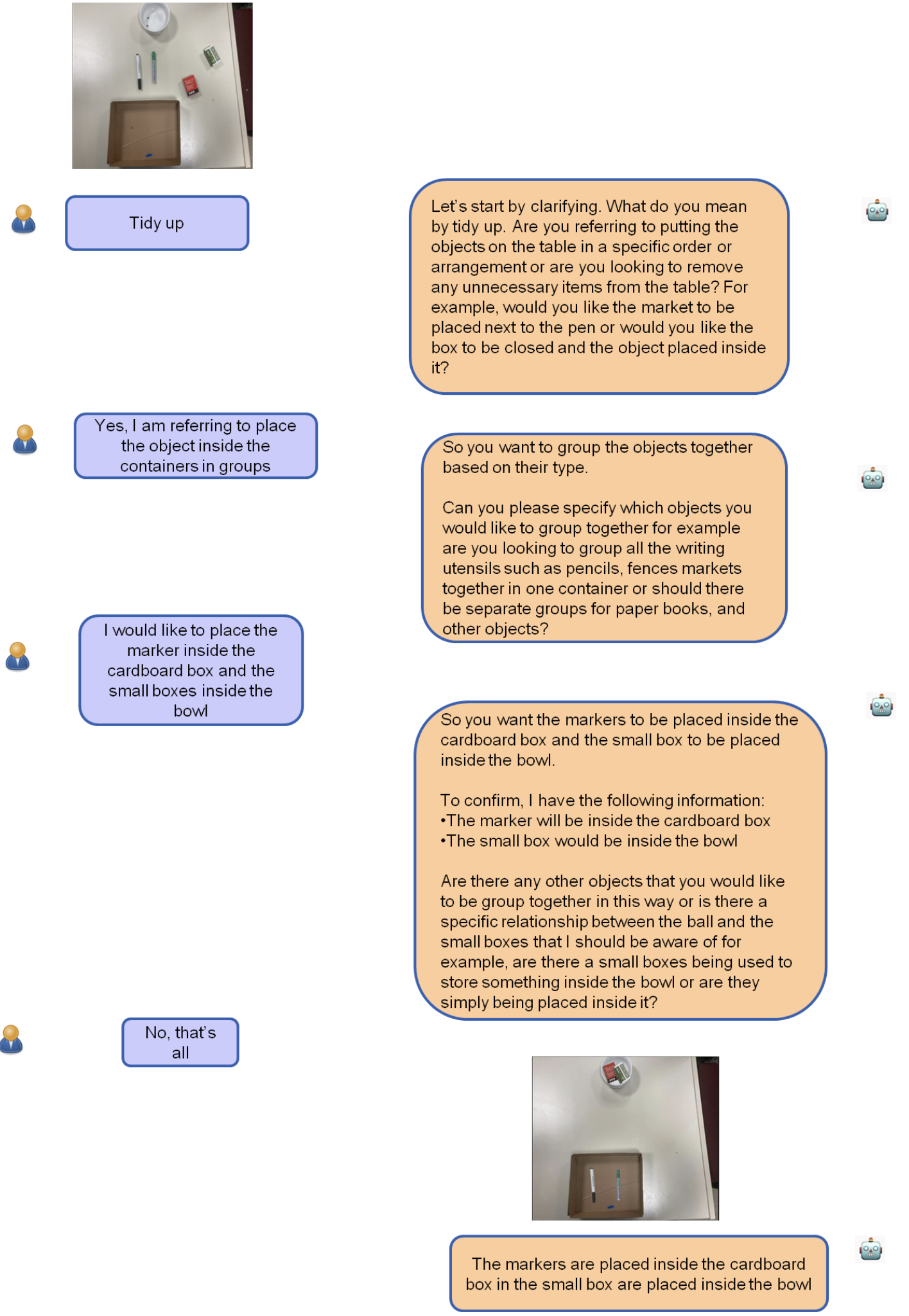}
    \caption{Example of clarification seeking behavior from CLARA. First the system seeks to clarify the abstract goal of tidying up and what exactly it refers to. Then, the system asks to specify how exactly the user wants the objects to be organized. When the dialog ends and CLARA has all the information needed, it generates the new scene and shows it to the user for verification.}
    \label{fig:chatting_example}
\end{figure}

\begin{table}[h]
\centering
\caption{Standard deviation and mean of cosine similarity for each model. CLARA achieves more consistent visual results when prompted with the same instruction.}
\label{tab:std}
\begin{tabular}{lcc}
\toprule
\textbf{System} & \textbf{stdev} & \textbf{cos} \\
\midrule
Gemini 2.0         & 0.1       & 0.92              \\
CLARA+Gemini 2.0   & 0.14       & 0.88              \\
CLARA              & \textbf{0.08}       & \textbf{0.95}        \\
\bottomrule
\end{tabular}
\end{table}

\begin{figure*}[t]
    \centering
    \includegraphics[width=1.0\textwidth]{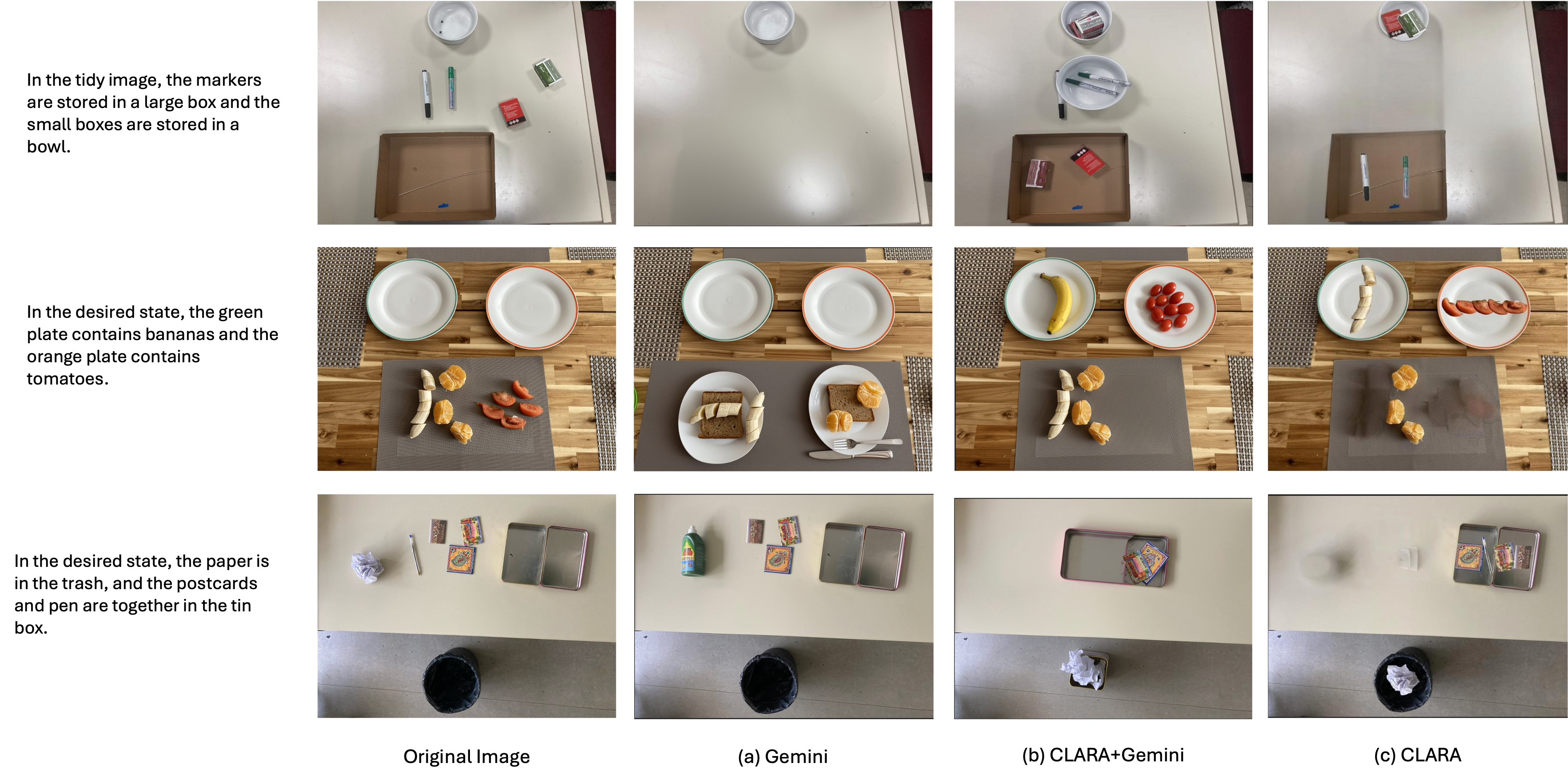}
    \caption{Visual outputs of each model in each of the abstract scenarios described: ``Organize a desk", ``Serve food to platters", and ``Clean a desk". As we can see, CLARA achieves a visual result which is aligned with the textual instruction while retaining the whole scene structure. Gemini either introduces new objects (such as new platters) or cannot capture the user's purpose (e.g. removing most objects on the table), since it does not perform clarification seeking behavior. CLARA+Gemini performs better than Gemini, but also introduces new objects on the scene (e.g. a new bowl) or erroneously modifies some objects.}
    \label{fig:image-variations}
\end{figure*}

The result consistency does indicate a measure of the system's understanding of the user's desires - however there could be cases where if the system does consistently something not asked by the user (e.g., just removing all objects from the scene), then the result is consistent, but not aligned with the user's purpose. Subsequently, we show qualitative results but also a user study to verify that our consistent results are also aligned with the human's goal.


\paragraph{Qualitative Results}
We show an example of chatting with CLARA in Figure~\ref{fig:chatting_example}.
Also, Figure \ref{fig:image-variations} shows sample outputs from Gemini along with outputs of CLARA for the same prompt, “Serve the food” Gemini’s outputs vary widely, sometimes placing objects in unrealistic positions or generating objects that were never requested.  In contrast, our pipeline yields a desired scene that closely follows the user’s specifications.

\subsection{User Study}
To further assess our system from a human perspective, and assess that it correctly aligns the visual result with the user's desires, we conducted a user study in order to compare our conversational grounding approach against Gemini.

\textbf{Participants and Procedure}
We recruited 6 participants, each participant interacted with all three systems (in random order) for the three scenarios (Tidying a Desk, Organizing a Floor, and Arranging a Kitchen). Participants then answered the following three questions using a 5-point Likert scale:

\begin{enumerate} \item \textit{"The final image accurately reflects my original intention."} \item \textit{"The system successfully retained the static scene components (i.e., the desk/tables were not changed visually)."} \item \textit{"I am satisfied with the interaction with the system."} \end{enumerate}

Participants’ responses (averaged) are shown in Table~\ref{tab:user_study}.

\begin{table}[h]
\centering
\caption{User Study Ratings (Mapped to 1--5 Likert Scale) for the three systems.}
\label{tab:user_study}
\begin{tabular}{lccc}
\toprule
\textbf{System} & \textbf{Q1 (Intent)} & \textbf{Q2 (Scene)} & \textbf{Q3 (Satisfaction)} \\
\midrule
Gemini 2.0         & 1.40       & 3.00       & 1.30       \\
CLARA+Gemini 2.0   & \textbf{2.97}       & 3.60       & 3.17       \\
CLARA              & 2.90       & \textbf{3.89}       & \textbf{3.37}       \\
\bottomrule
\end{tabular}
\end{table}

As shown in Table~\ref{tab:user_study}, the standard Gemini 2.0 approach (lacking clarification-seeking behavior) yields notably lower scores for both user intent fulfillment and overall satisfaction. By contrast, incorporating clarification-seeking (i.e., the CLARA\* systems) significantly improves these metrics. Moreover, regarding static or unmovable scene elements, the CLARA\* systems demonstrate a stronger capability to preserve the underlying structure of the scene while repositioning objects as specified. Interestingly, passing the extracted directives back into Gemini sometimes led to unsatisfactory outcomes; however, in certain cases where the directives were sufficiently abstract for our image-editing module to parse, Gemini’s generative process performed better.

\subsection{Discussion}

Our experiments highlight the value of clarification-seeking behavior in large vision-language models for abstract/vague commands. LVLMs when left without specific guidance tend to overestimate their capabilities in grounding vague commands, resulting in many cases in drastically different outcomes compared to the desires or the instructor. As shown both in our qualitative results, as well as our user study and quantitative metrics, the proposed system tends to generate both more consistent results (as demonstrated by Table~\ref{tab:std}), but also results that are aligned with the user's intensions, as demonstrated by our user study. Moving forward it is important to instill such kind of clarification seeking behavior more implicitly in such models during training, imitating how initially children tend to overestimate their understanding, while adults engage in dialogue to further establish common ground. 

\section{Conclusion}
This paper presented CLARA, a cognitively-inspired framework that proactively resolves ambiguous directives through targeted clarification questions. By modeling the clarification-seeking process inherent in human communication and development, we demonstrated that CLARA achieves more reliable visual outcomes, aligned with the user's intention In the future, we plan to extend CLARA’s capabilities to dynamic robot operations in real-world environments. One promising direction is to integrate the final target image into a reinforcement learning loop, using visual similarity or arrangement correctness as a reward mechanism. 


\bibliographystyle{IEEEtran}
\bibliography{references}

\end{document}